# A Comprehensive Survey of Data Mining-based Fraud Detection Research

CLIFTON PHUA[1*], VINCENT LEE[1], KATE SMITH[1] & ROSS GAYLER[2]

[1]School of Business Systems, Faculty of Information Technology, Monash University, Clayton campus, Wellington Road, Clayton, Victoria 3800, Australia (*author for correspondence, tel.: +61-3-99052065)
{e-mail: clifton.phua, vincent.lee, kate.smith@infotech.monash.edu.au};

[2]Baycorp Advantage, Level 12, 628 Bourke Street, Melbourne, VIC 3000, Australia
{e-mail: r.gayler@mbox.com.au}

## ABSTRACT

This survey paper categorises, compares, and summarises from almost all published technical and review articles in automated fraud detection within the last 10 years. It defines the professional fraudster, formalises the main types and subtypes of known fraud, and presents the nature of data evidence collected within affected industries. Within the business context of mining the data to achieve higher cost savings, this research presents methods and techniques together with their problems. Compared to all related reviews on fraud detection, this survey covers much more technical articles and is the only one, to the best of our knowledge, which proposes alternative data and solutions from related domains.

## 1. INTRODUCTION & MOTIVATION

Data mining is about finding insights which are statistically reliable, unknown previously, and actionable from data (Elkan, 2001). This data must be available, relevant, adequate, and clean. Also, the data mining problem must be well-defined, cannot be solved by query and reporting tools, and guided by a data mining process model (Lavrac *et al*, 2004).

The term fraud here refers to the abuse of a profit organisation's system without necessarily leading to direct legal consequences. In a competitive environment, fraud can become a business critical problem if it is very prevalent and if the prevention procedures are not fail-safe. Fraud detection, being part of the overall fraud control, automates and helps reduce the manual parts of a screening/checking process. This area has become one of the most established industry/government data mining applications.

It is impossible to be absolutely certain about the legitimacy of and intention behind an application or transaction. Given the reality, the best cost effective option is to tease out possible evidences of fraud from the available data using mathematical algorithms.

Evolved from numerous research communities, especially those from developed countries, the analytical engine within these solutions and software are driven by artificial immune systems, artificial intelligence, auditing, database, distributed and parallel computing, econometrics, expert systems, fuzzy logic, genetic algorithms, machine learning, neural networks, pattern recognition, statistics, visualisation and others. There are plenty of specialised fraud detection solutions and software[1] which protect businesses such as credit card, e-commerce, insurance, retail, telecommunications industries.

There are often two main criticisms of data mining-based fraud detection research: the dearth of publicly available real data to perform experiments on; and the lack of published well-researched methods and techniques. To counter both of them, this paper garners all related literature for categorisation and comparison, selects some innovative methods and techniques for discussion; and points toward other data sources as possible alternatives.

- The primary objective of this paper is to define existing challenges in this domain for the different types of large data sets and streams. It categorises, compares, and summarises relevant data mining-based fraud detection methods and techniques in published academic and industrial research.

- The second objective is to highlight promising new directions from related adversarial data mining fields/applications such as epidemic/outbreak detection, insider trading, intrusion detection, money laundering, spam detection, and terrorist detection. Knowledge and experience from these adversarial domains can be interchangeable and will help prevent repetitions of common mistakes and "reinventions of the wheel".

Section 2 – Who are the white collar criminals which a fraud detection system should be designed to discover? Where can one apply data mining techniques to commercial fraud? Section 3 – What data is available for fraud detection? Which performance measurements are appropriate for analysis? Section 4 – Which techniques often used for automated fraud detection? What combinations of techniques have been recommended? What are their weaknesses? Sections 5 – What analytical methods and techniques from other adversarial domains can one apply in fraud detection? Section 6 – How is this fraud detection survey different from others? Section 7 concludes with a brief summary.

## 2. BACKGROUND
This section highlights the types of fraudsters and affected industries.

### 2.1 Fraudsters

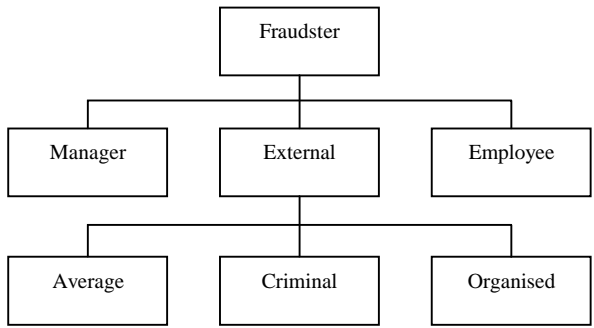

**Figure 2.1:** Hierarchy chart of white-collar crime perpetrators from both firm-level and community-level perspectives.

With reference to figure 2.1, the profit-motivated fraudster has interactions with the affected business. Traditionally, each business is always susceptible to internal fraud or corruption from its management (high-level) and non-management employees (low-level). In addition to internal and external audits for fraud control, data mining can also be utilised as an analytical tool.

From figure 1, the fraudster can be an external party, or parties. Also, the fraudster can either commit fraud in the form of a prospective/existing customer (consumer) or a prospective/existing supplier (provider). The external fraudster has three basic profiles: the average offender, criminal offender, and organised crime offender. Average offenders display random and/or occasional dishonest behaviour when there is opportunity, sudden temptation, or when suffering from financial hardship.

In contrast, the more risky external fraudsters are individual criminal offenders and organised/group crime offenders (professional/career fraudsters) because they repeatedly disguise their true identities and/or evolve their *modus operandi* over time to approximate legal forms and to counter detection systems. Therefore, it is important to account for the strategic interaction, or moves and countermoves, between a fraud detection system's algorithms and the professional fraudsters' *modus operandi*. It is probable that internal and insurance fraud is more likely to be committed by average offenders; credit and telecommunications fraud is more vulnerable to professional fraudsters.

For many companies where they have interactions with up to millions of external parties, it is cost-prohibitive to manually check the majority of the external parties' identities and activities. So the riskiest ones determined through data mining output such as suspicion scores, rules, and visual anomalies will be investigated.

### 2.2 Affected Commercial Industries

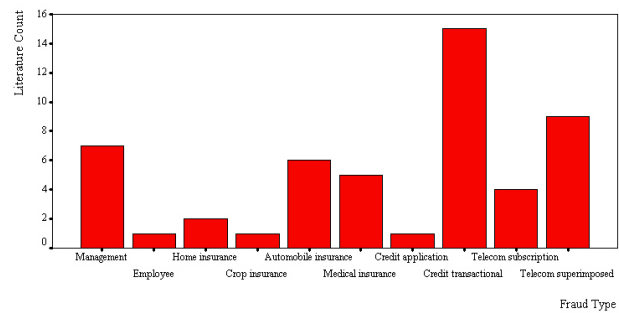

**Figure 2.2:** Bar chart of fraud types from 51 unique and published fraud detection papers. The most recent publication is used to represent previous similar publications by the same author(s).

Figure 2.2 details the subgroups of internal, insurance, credit card, and telecommunications fraud detection. Internal fraud detection is concerned with determining fraudulent financial reporting by management (Lin *et al*, 2003; Bell and Carcello, 2000; Fanning and Cogger, 1998; Summers and Sweeney, 1998; Beneish, 1997; Green and Choi, 1997), and abnormal retail transactions by employees (Kim *et al*, 2003). There are four subgroups of insurance fraud detection: home insurance (Bentley, 2000; Von Altrock, 1997), crop insurance (Little *et al*, 2002), automobile insurance (Phua *et al*, 2004; Viaene *et al*, 2004; Brockett *et al*, 2002; Stefano and Gisella, 2001; Belhadji *et al*, 2000; Artis *et al*, 1999), and medical insurance (Yamanishi *et al*, 2004; Major and Riedinger, 2002; Williams, 1999; He *et al*, 1999; Cox, 1995). Credit fraud detection refers to screening credit applications (Wheeler and Aitken, 2000), and/or logged credit card transactions (Fan, 2004; Chen *et al*, 2004; Chiu and Tsai, 2004; Foster and Stine, 2004; Kim and Kim, 2002; Maes *et al*, 2002; Syeda *et al*, 2002; Bolton and Hand, 2001; Bentley *et al*, 2000; Brause *et al*, 1999; Chan *et al*, 1999; Aleskerov *et al*, 1997; Dorronsoro *et al*, 1997; Kokkinaki, 1997; Ghosh and Reilly, 1994). Similar to credit fraud detection, telecommunications subscription data (Cortes *et al*, 2003; Cahill *et al*, 2002; Moreau and Vandewalle, 1997; Rosset *et al*, 1999), and/or wire-line and wire-less phone calls (Kim *et al*, 2003; Burge and Shawe-Taylor, 2001; Fawcett and Provost, 1997; Hollmen and Tresp, 1998; Moreau *et al*, 1999; Murad and Pinkas, 1999; Taniguchi *et al*, 1998; Cox, 1997; Ezawa and Norton, 1996) are monitored.

Credit transactional fraud detection has received the most attention from researchers although it has been loosely used here to include bankruptcy prediction (Foster and Stine, 2004) and bad debts prediction (Ezawa and Norton, 1996). Employee/retail (Kim *et al*, 2003), national crop insurance (Little *et al*, 2002), and credit application (Wheeler and Aitken, 2000) each has only one academic publication.

The main purpose of these detection systems is to identify general trends of suspicious/fraudulent applications and transactions. In the case of application fraud, these fraudsters apply for insurance entitlements using falsified information, and apply for credit and telecommunications products/services using non-existent identity information or someone else's identity information. In the case of transactional fraud, these fraudsters take over or add to the usage of an existing legitimate credit or telecommunications account.

There are other fraud detection domains. E-businesses and e-commerce on the Internet present a challenging data mining task because it blurs the boundaries between fraud detection systems and network intrusion detection systems. Related literature focus on video-on-demand websites (Barse *et al*, 2003) and IP-based telecommunication services (McGibney and Hearne, 2003). Online sellers (Bhargava *et al*, 2003) and online buyers (Sherman, 2002) can be monitored by automated systems. Fraud detection in government organisations such as tax (Bonchi *et al*, 1999) and customs (Shao *et al*, 2002) has also been reported.

## 3. DATA AND MEASUREMENTS

This section discusses the types of available data and previously used performance measures.

### 3.1 Structured data

This subsection aims to define the attributes and examples which have been used for previous fraud detection experimental studies and actual systems. By doing so, future studies on fraud detection will find this useful to either validate their real data or create synthetic data.

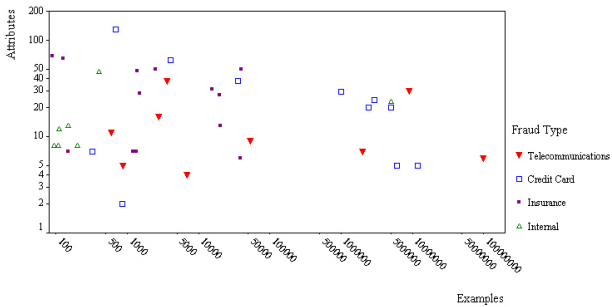

**Figure 3.1:** Scatter plot of the data size from 40 unique and published fraud detection papers within common fraud types.

Figure 3.1 shows the number of original attributes (vertical axis) and pre-sampled examples (horizontal axis) from internal, insurance, credit card, and telecommunications fraud detection literature.

Generally, attributes can be binary, numerical (interval or ratio scales), categorical (nominal or ordinal scales), or a mixture of the three. 16 data sets have less than 10 attributes, 18 data sets have between 10 to 49 attributes, 5 data sets have between 50 to 99 attributes, and only 1 data set used more than 100 attributes (Wheeler and Aitken, 2000).

Management data sets are the smallest (all have less than 500 examples), except for employee/retail data with more than 5 million transactions (Kim *et al*, 2003). Insurance data sets consist of hundreds of examples and the largest contain 40000 examples (Williams, 1999). Most credit transactional data have more than 1 million transactions and the largest contain more than 12 million transactions per year (Dorronsoro *et al*, 1997). Telecommunications data are the largest because they comprise of transactions generated by hundreds, thousands, or millions of accounts. The largest reported is produced by at least 100 million telecommunications accounts (Cortes *et al*, 2003).

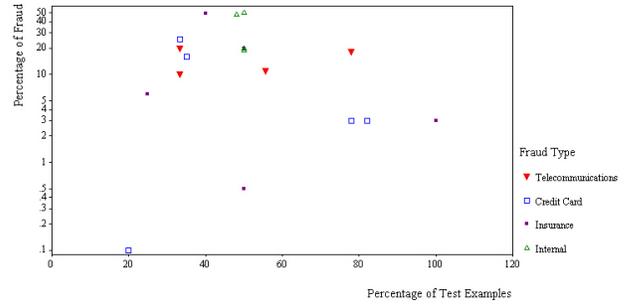

**Figure 3.2:** Scatter plot of the percentage of fraud and percentage of test of entire data set. 19 unique and published fraud detection papers within common fraud types were used.

Figure 3.2 shows the percentage of fraud (vertical axis) and percentage of test examples (horizontal axis) of the entire data set described in each study.

Six studies using credit transactional and insurance data have less than 10 percent fraud. In particular, Foster and Stine (2004) and Bentley (2000) have as low as 0.1 percent fraud in credit transactional data and 0.5 percent fraud in home insurance data respectively. More than 80 percent (16 papers) of the 19 papers has skewed data with less than 30% fraud. The average of the proportion of test examples to total examples of the 19 papers is around 50%.

The specific attributes used for detecting each fraud type are generally the same. Management data are typically financial ratios using accounts receivable, allowance of doubtful debts, and net sales figures. Crop insurance data consist of ratios using amount of compensation, premium, and liability figures (Little *et al*, 2002). Home insurance data is made up of customer behaviour (current claim amount, time as customer, past claims) and financial status (annual income, average bank balance, number of overdrafts) (Von Altrock, 1995). Automobile insurance data are usually binary indicators grouped into accident, claimant, driver, injury, treatment, lost wages, vehicle, and other categories. Medical insurance data can comprise of patient demographics (age and gender), treatment details (services), and policy and claim details (benefits and amount) (Williams, 1999).

Specific attributes in credit transaction data are often not revealed but they should comprise of date/time stamps, current transaction (amount, geographical location, merchant industry code and validity code), transactional history, payment history, and other account information (age of account) (Chan *et al*, 1999; Ghosh and Reilly, 1994).

Telecommunications data can comprise of individual call information (date/time stamps, source number, destination number, call duration, type of call, geographical origin, geographical destination) (Cortes *et al*, 2003; Fawcett and Provost, 1997) and account summary information (likely payment methods, average monthly bill, average time between calls, daily and weekly summaries of individual calls) (Cahill *et al*, 2002; Taniguchi *et al*, 1998).

Apart from date/time attributes in Kim *et al* (2003) on employee/retail fraud and day/week/month/year attributes in Phua *et al* (2004) on automobile insurance fraud, the data collected to detect internal and insurance fraud are static and do not provide any temporal information. Not only does insurance (pathology provider) data described in Yamanishi *et al* (2004) lack temporal information, the other attributes such as proportion of tests performed are not effective for fraud detection.

Almost all the data has been de-identified, apart from Wheeler and Aitken (2000) which describes the use of identity information such as names and addresses from credit applications. While most telecommunication account data are behavioural, Rosset *et al* (1999) includes de-identified demographic data such as age and ethnicity for the telecommunications customer.

There are no publicly available data sets for studying fraud detection, except for a relatively small automobile insurance data set used in Phua *et al* (2004). And obtaining real data from companies for research purposes is extremely hard due to legal and competitive reasons. To circumvent these data availability problems and work on a particular fraud type, one alternative is to create synthetic data which matches closely to actual data. Barse *et al* (2003) justifies that synthetic data can train and adapt a system without any data on known frauds, variations of known fraud and new frauds can be artificially created, and to benchmark different systems. In addition, they summarised important qualities which should be in simulated data and proposed a five-step synthetic data generation methodology.

Barse *et al* (2003) reported that use of simulated data had mixed results when applied to real data. Three out of the 51 papers presented in Figure 2.2 used simulated data but the credit transaction data was either not realistic (Chen *et al*, 2004; Aleskerov *et al*, 1997) or the insurance data and results were not explained (Pathak *et al*, 2003).

The next alternative, according to Fawcett (2003), is to mine email data for spam because researchers can study many of the same data issues as fraud detection and the spam data is available publicly in large quantities. In contrast to the structured data collected for fraud detection, unstructured email data will require effective feature selection or text processing operations.

## 3.2 Performance Measures

Most fraud departments place monetary value on predictions to maximise cost savings/profit and according to their policies. They can either define explicit cost (Phua *et al*, 2004; Chan *et al*, 1999; Fawcett and Provost, 1997) or benefit models (Fan *et al*, 2004; Wang *et al*, 2003).

Cahill *et al* (2002) suggests giving a score for an instance (phone call) by determining the similarity of it to known fraud examples (fraud styles) divided by the dissimilarity of it to known legal examples (legitimate telecommunications account).

Most of the fraud detection studies using supervised algorithms since 2001 have abandoned measurements such as true positive rate (correctly detected fraud divided by actual fraud) and accuracy at a chosen threshold (number of instances predicted correctly, divided by the total number of instances). In fraud detection, misclassification costs (false positive and false negative error costs) are unequal, uncertain, can differ from example to example, and can change over time. In fraud detection, a false negative error is usually more costly than a false positive error. Regrettably, some recent studies on credit card transactional fraud (Chen *et al*, 2004) and telecommunications superimposed fraud (Kim *et al*, 2003) still aim to only maximise accuracy. Some use Receiver Operating Characteristic (ROC) analysis (true positive rate versus false positive rate).

Apart from Viaene *et al* (2004), no other fraud detection study on supervised algorithms has sought to maximise Area under the Receiver Operating Curve (AUC) and minimise cross entropy (CXE). AUC measures how many times the instances have to be swapped with their neighbours when sorting data by predicted scores; and CXE measures how close predicted scores are to target scores. In addition, Viaene *et al* (2004) and Foster and Stine (2004) seek to minimise Brier score (mean squared error of predictions). Caruana and Niculescu-Mizil (2004) argues that the most effective way to assess supervised algorithms is to use one metric from threshold, ordering, and probability metrics; and they justify using the average of mean squared error, accuracy, and AUC. Fawcett and Provost (1999) recommend Activity Monitoring Operating Characteristic (AMOC) (average score versus false alarm rate) suited for timely credit transactional and telecommunications superimposition fraud detection.

For semi-supervised approaches such as anomaly detection, Lee and Xiang (2001) propose entropy, conditional entropy, relative conditional entropy, information gain, and information cost. For unsupervised algorithms, Yamanishi *et al* (2004) used the Hellinger and logarithmic scores to find statistical outliers for insurance; Burge and Shawe-Taylor (2001) employed Hellinger score to determine the difference between short-term and long-term profiles for the telecommunications account. Bolton and Hand (2001) recommends the *t*-statistic as a score to compute the standardised distance of the target account with centroid of the peer group; and also to detect large spending changes within accounts.

Other important considerations include how fast the frauds can be detected (detection time/time to alarm), how many styles/types of fraud detected, whether the detection was done in online/real time (event-driven) or batch mode (time-driven) (Ghosh and Reilly, 1994).

There are problem-domain specific criteria in insurance fraud detection. To evaluate automated insurance fraud detection, some domain expert comparisons and involvement have been described. Von Altrock (1995) claimed that their algorithm performed marginally better than the experienced auditors. Brockett *et al* (2002) and Stefano and Gisella (2001) summed up their performance as being consistent with the human experts and their regression scores. Belhadji *et al* (2000) stated that both automated and manual methods are complementary. Williams (1999) supports the role of the fraud specialist to explore and evolve rules. NetMap (2004) reports visual analysis of insurance claims by the user helped discover the fraudster.

# 4. METHODS AND TECHNIQUES

This section examines four major methods commonly used, and their corresponding techniques and algorithms.

## 4.1 Overview

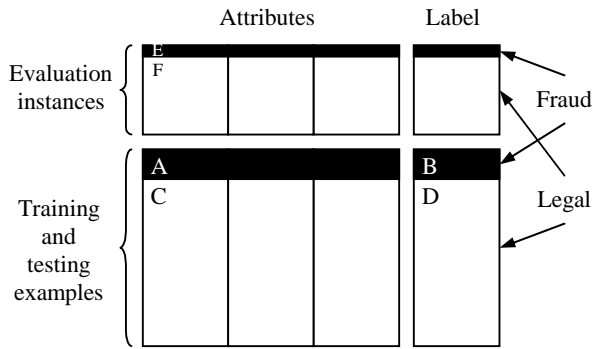

**Figure 4.1:** Structured diagram of the possible data for analysis. Data mining approaches can utilise training/testing data with labels, only legal examples, and no labels to predict/describe the evaluation data.

Figure 4.1 shows that many existing fraud detection systems typically operate by adding fraudulent claims/applications/transactions/accounts/sequences (A) to "black lists" to match for likely frauds in the new instances (E). Some use hard-coded rules which each transaction should meet such as matching addresses and phone numbers, and price and amount limits (Sherman, 2002).

An interesting idea borrowed from spam (Fawcett, 2003, p144, figure 5) is to understand the temporal nature of fraud in the "black lists" by tracking the frequency of terms and category of terms (style or strategy of fraudster) found in the attributes of fraudulent examples over time. Below outlines the complex nature of data used for fraud detection in general (Fawcett, 2003; 1997):

- Volume of both fraud and legal classes will fluctuate independently of each other; therefore class distributions (proportion of illegitimate examples to legitimate examples) will change over time.
- Multiple styles of fraud can happen at around the same time. Each style can have a regular, occasional, seasonal, or once-off temporal characteristic.
- Legal characteristics/behaviour can change over time.
- Within the near future after uncovering the current *modus operandi* of professional fraudsters, these same fraudsters will continually supply new or modified styles of fraud until the detection systems start generating false negatives again.

With reference to Figure 4.1, the common data mining approaches to determine the most suspicious examples from the incoming data stream (evaluation data) are:

1) Labelled training data (A + B + C + D) can be processed by single *supervised* algorithms (Section 4.2). A better suggestion is to employ hybrids such as multiple supervised algorithms (Section 4.3.1), or both supervised and unsupervised algorithms (Section 4.3.2) to output suspicion scores, rules and/or visual anomalies on evaluation data.

2) All known legal claims/applications/transactions/accounts/sequences (C) should be used processed by *semi-supervised* algorithms to detect significant anomalies from consistent normal behaviour (Section 4.4).

However, there are many criticisms with using labelled data to detect fraud:

- In an operational event-driven environment, the efficiency of processing is critical.
- The length of time needed to flag examples as fraudulent will be the same amount of time the new fraud types will go unnoticed.
- The class labels of the training data can be incorrect and subject to sample selectivity bias (Hand, 2004).
- They can be quite expensive and difficult to obtain (Brockett, 2002).
- Staffs have to manually label each example and this has the potential of breaching privacy particularly if the data contains identity and personal information.
- Dorronsoro *et al* (1997) recommend the use of *unlabelled* data because the fraudster will try to make fraud and legal classes hard to distinguish.

Therefore it is necessary to:

3) Combine training data (the class labels are not required here) with evaluation data (A + C + E + F). These should be processed by single or multiple *unsupervised* algorithms to output suspicion scores, rules and/or visual anomalies on evaluation data (Section 4.5).

## 4.2 Supervised Approaches on Labelled Data (A + B + C + D)

Predictive supervised algorithms examine all previous labelled transactions to mathematically determine how a standard fraudulent transaction looks like by assigning a risk score (Sherman, 2002). Neural networks are popular and support vector machines (SVMs) have been applied. Ghosh and Reilly (1994) used a three-layer, feed-forward Radial Basis Function (RBF) neural network with only two training passes needed to produce a fraud score in every two hours for new credit card transactions. Barse *et al* (2003) used a multi-layer neural network with exponential trace memory to handle temporal dependencies in synthetic Video-on-Demand log data. Syeda *et al* (2002) propose fuzzy neural networks on parallel machines to speed up rule production for customer-specific credit card fraud detection. Kim *et al* (2003) proposes SVM ensembles with either bagging and boosting with aggregation methods for telecommunications subscription fraud.

The neural network and Bayesian network comparison study (Maes *et al*, 2002) uses the STAGE algorithm for Bayesian networks and backpropagation algorithm for neural networks in credit transactional fraud detection. Comparative results show that Bayesian networks were more accurate and much faster to train, but Bayesian networks are slower when applied to new instances.

Ezawa and Norton (1996) developed Bayesian network models in four stages with two parameters. They argue that regression, nearest-neighbour, and neural networks are too slow and decision trees have difficulties with certain discrete variables. The model with most variables and with some dependencies performed best for their telecommunications uncollectible debt data.

Viaene *et al* (2004) applies the weight of evidence formulation of AdaBoosted naive Bayes (boosted fully independent Bayesian network) scoring. This allows the computing of the relative importance (weight) for individual components of suspicion and displaying the aggregation of evidence pro and contra fraud as a balance of evidence which is governed by a simple additivity principle. Compared to unboosted and boosted naive Bayes, the framework showed slightly better accuracy and AUC but clearly improved on the cross entropy and Brier scores. It is also readily accessible and naturally interpretable decision support and allows for flexible human expert interaction and tuning on an automobile insurance dataset.

Decision trees, rule induction, and case-based reasoning have also been used. Fan (2004) introduced systematic data selection to mine concept-drifting, possibly insufficient, data streams. The paper proposed a framework to select the optimal model from four different models (based on old data chunk only, new data chunk only, new data chunk with selected old data, and old and new data chunks). The selected old data is the examples which both optimal models at the consecutive time steps predict correctly. The cross-validated decision tree ensemble is consistently better than all other decision tree classifiers and weighted averaging ensembles under all concept-drifting data chunk sizes, especially when the new data chunk size of the credit card transactions are small. With the same credit card data as Fan (2004), Wang *et al* (2003) demonstrates a pruned classifier C4.5 ensemble which is derived by weighting each base classifier according to its expected benefits and then averaging their outputs. The authors show that the ensemble will most likely perform better than a single classifier which uses exponential weighted average to emphasise more influence on recent data.

Rosset *et al* (1999) presents a two-stage rules-based fraud detection system which first involves generating rules using a modified C4.5 algorithm. Next, it involves sorting rules based on accuracy of customer level rules, and selecting rules based on coverage of fraud of customer rules and difference between behavioural level rules. It was applied to a telecommunications subscription fraud. Bonchi *et al* (1999) used boosted C5.0 algorithm on tax declarations of companies. Shao *et al* (2002) applied a variant of C4.5 for customs fraud detection.

Case-based reasoning (CBR) was used by Wheeler and Aitken (2000) to analyse the hardest cases which have been misclassified by existing methods and techniques. Retrieval was performed by thresholded nearest neighbour matching. Diagnosis utilised multiple selection criteria (probabilistic curve, best match, negative selection, density selection, and default) and resolution strategies (sequential resolution-default, best guess, and combined confidence) which analysed the retrieved cases. The authors claimed that CBR had 20% higher true positive and true negative rates than common algorithms on credit applications.

Statistical modelling such as regression has been extensively utilised. Foster and Stine (2004) use least squares regression and stepwise selection of predictors to show that standard statistical methods are competitive. Their version of fully automatic stepwise regression has three useful modifications: firstly, organises calculations to accommodate interactions; secondly, exploits modern decision-theoretic criteria to choose predictors; thirdly, conservatively estimate p-values to handle sparse data and a binary response before calibrating regression predictions. If cost of false negative is much higher than a false positive, their regression model obtained significantly lesser misclassification costs than C4.5 for telecommunications bankruptcy prediction.

Belhadji *et al* (2000) chooses the best indicators (attributes) of fraud by first querying domain experts, second calculating conditional probabilities of fraud for each indicator and third Probit regressions to determine most significant indicators. The authors also use Prohit regressions to predict fraud and adjusts the threshold to suit company fraud policy on automobile property damages. Artis *et al* (1999) compares a multinomial logit model (MNL) and nested multinomial logit model (NMNL) on a multiclass classification problem. Both models provide estimated conditional probabilities for the three classes but NMNL uses the two step estimation for its nested choice decision tree. It was applied to automobile insurance data. Mercer (1990) described least-squares stepwise regression analysis for anomaly detection on aggregated employee's applications data.

Other techniques include expert systems, association rules, and genetic programming. Expert systems have been applied to insurance fraud. Major and Riedinger (2002) have implemented an actual five-layer expert system in which expert knowledge is integrated with statistical information assessment to identify medical insurance fraud. Pathak *et al* (2003), Stefano and Gisella (2001) and Von Altrock (1997) have experimented on fuzzy expert systems. Deshmukh and Talluru (1997) applied an expert system to management fraud. Chiu and Tsai (2004) introduce a Fraud Patterns Mining (FPM) algorithm, modified from *Apriori*, to mine a common format for fraud-only credit card data. Bentley (2000) uses genetic programming with fuzzy logic to create rules for classifying data. This system was tested on real home insurance claims (Bentley, 2000) and credit card transaction data (Bentley *et al*, 2000). None of these papers on expert systems, association rules, and genetic programming provide any direct comparisons with the many other available methods and techniques.

The above supervised algorithms are conventional learning techniques which can only process structured data from single 1-to-1 data tables. Further research using labelled data in fraud detection can benefit from applying relational learning approaches such as Inductive Logic Programming (ILP) (Muggleton and DeRaedt, 1994) and simple homophily-based classifiers (Provost *et al*, 2003) on relational databases. Perlich and Provost (2003) also present novel target-dependent aggregation methods for converting the relational learning problem into a conventional one.

## 4.3 Hybrid Approaches with Labelled Data

### 4.3.1 *Supervised Hybrids (A + B + C + D)*

Popular supervised algorithms such as neural networks, Bayesian networks, and decision trees have been combined or applied in a sequential fashion to improve results. Chan *et al* (1999) utilises naive Bayes, C4.5, CART, and RIPPER as base classifiers and stacking to combine them. They also examine bridging incompatible data sets from different companies and the pruning of base classifiers. The results indicate high cost savings and better efficiency on credit card transactions. Phua *et al* (2004) proposes backpropagation neural networks, naive Bayes, and C4.5 as base classifiers on data partitions derived from minority oversampling with replacement. Its originality lies in the use of a single meta-classifier (stacking) to choose the best base classifiers, and then combine these base classifiers' predictions (bagging) to produce the best cost savings on automobile insurance claims.

Ormerod *et al* (2003) recommends a rule generator to refine the weights of the Bayesian network. Kim and Kim (2002) propose a decision tree to partition the input space, tanh as a weighting function to generate fraud density, and subsequently a backpropagation neural network to generate a weighted suspicion score on credit card transactions.

Also, He *et al* (1999) propose genetic algorithms to determine optimal weights of the attributes, followed by *k*-nearest neighbour algorithm to classify the general practitioner data. They claim significantly better results than without feature weights and when compared to CBR.

### 4.3.2 *Supervised/Unsupervised Hybrids (A + B + C + D)*

There is extensive work on labelled data using both supervised and unsupervised algorithms in telecommunications fraud detection. Cortes and Pregibon (2001) propose the use of signatures (telecommunication account summaries) which are updated daily (time-driven). Fraudulent signatures are added to the training set and processed by supervised algorithms such as atree, slipper, and model-averaged regression. The authors remark that fraudulent toll-free numbers tend to have extensive late night activity and long call durations. Cortes and Pregibon (2001) use signatures assumed to be legitimate to detect significant changes in calling behaviour. Association rules is used to discover interesting country combinations and temporal information from the previous month. A graph-theoretic method (Cortes *et al*, 2003) is used to visually detect communities of interest of fraudulent international call accounts (see Section 4.5). Cahill *et al* (2002) assign an averaged suspicion score to each call (event-driven) based on its similarity to fraudulent signatures and dissimilarity to its account's normal signature. Calls with low scores are used to update the signature and recent calls are weighted more heavily than earlier ones in the signature.

Fawcett and Provost (1997) present fraud rule generation from each cloned phone account's labelled data and rule selection to cover most accounts. Each selected fraud rule is applied in the form of monitors (number and duration of calls) to the daily legitimate usage of each account to find anomalies. The selected monitors' output and labels on an account's previous daily behaviour are used as training data for a simple Linear Threshold Unit. An alarm will be raised on that account if the suspicion score on the next evaluation day exceeds its threshold. In terms of cost savings and accuracy, this method performed better than other methods such as expert systems, classifiers trained without account context, high usage, collision detection, velocity checking, and dialled digit analysis on detecting telecommunications superimposed fraud.

Two studies on telecommunications data show that supervised approaches achieve better results than unsupervised ones. With AUC as the performance measure, Moreau *et al* (1999) show that supervised neural network and rule induction algorithms outperform two forms of unsupervised neural networks which identify differences between short-term and long-term statistical account behaviour profiles. The best results are from a hybrid model which combines these four techniques using logistic regression. Using true positive rate with no false positives as the performance measure, Taniguchi *et al* (1998) claim that supervised neural networks and Bayesian networks on labelled data achieve significantly better outcomes than unsupervised techniques such as Gaussian mixture models on each non-fraud user to detect anomalous phone calls.

Unsupervised approaches have been used to segment the insurance data into clusters for supervised approaches. Williams and Huang (1997) applies a three step process: *k*-means for cluster detection, C4.5 for decision tree rule induction, and domain knowledge, statistical summaries and visualisation tools for rule evaluation. Williams (1999) use a genetic algorithm, instead of C4.5, to generate rules and to allow the domain user, such as a fraud specialist, to explore the rules and to allow them to evolve accordingly on medical insurance claims. Brockett *et al* (1998) present a similar methodology utilising the Self Organising Maps (SOM) for cluster detection before backpropagation neural networks in automobile injury claims. Cox (1995) uses an unsupervised neural network followed by a neuro-fuzzy classification system to monitor medical providers' claims.

Unconventional hybrids include the use of backpropagation neural networks, followed by SOMs to analyse the classification results on medical providers' claims (He *et al*, 1997) and RBF neural networks to check the results of association rules for credit card transactions (Brause *et al*, 1999).

## 4.4 Semi-supervised Approaches with Only Legal (Non-fraud) Data (C)

Kim *et al* (2003) implements a novel fraud detection method in five steps: First, generate rules randomly using association rules algorithm *Apriori* and increase diversity by a calender schema; second, apply rules on known legitimate transaction database, discard any rule which matches this data; third, use remaining rules to monitor actual system, discard any rule which detects no anomalies; fourth, replicate any rule which detects anomalies by adding tiny random mutations; and fifth, retain the successful rules. This system has been and currently being tested for internal fraud by employees within the retail transaction processing system.

Murad and Pinkas (1999) use profiling at call, daily, and overall levels of normal behaviour from each telecommunications account. The common daily profiles are extracted using a clustering algorithm with cumulative distribution distance function. An alert is raised if the daily profile's call duration, destination, and quantity exceed the threshold and standard deviation of the overall profile. Aleskerov *et al* (1997) experiment with auto-associative neural networks (one hidden layer and the same number of input and output neurons) on each credit card account's legal transactions. Kokkinaki (1997) proposes similarity trees (decision trees with Boolean logic functions) to profile each legitimate customer's behaviour to detect deviations from the norm and cluster analysis to segregate each legitimate customer's credit card transactions.

## 4.5 Unsupervised Approaches with Unlabelled Data (A + C + E + F)

Link analysis and graph mining are hot research topics in anti-terrorism, law enforcement, and other security areas, but these techniques seem to be relatively under-rated in fraud detection research. A white paper (NetMap, 2004) describes how the emergent group algorithm is used to form groups of tightly connected data and how it led to the capture of an actual elusive fraudster by visually analysing twelve months worth of insurance claims. There is a brief application description of a visual telecommunications fraud detection system (Cox, 1997) which flexibly encodes data using colour, position, size and other visual characteristics with multiple different views and levels. The intuition is to combine human detection with machine computation.

Cortes *et al* (2001) examines temporal evolution of large dynamic graphs' for telecommunications fraud detection. Each graph is made up of subgraphs called Communities Of Interest (COI). To overcome instability of using just the current graph, and storage and weightage problems of using all graphs at all time steps; the authors used the exponential weighted average approach to update subgraphs daily. By linking mobile phone accounts using call quantity and durations to form COIs, the authors confirm two distinctive characteristics of fraudsters. First, fraudulent phone accounts are linked - fraudsters call each other or the same phone numbers. Second, fraudulent call behaviour from flagged frauds are reflected in some new phone accounts - fraudsters retaliate with application fraud/identity crime after being detected. Cortes *et al* (2003) states their contribution to dynamic graph research in the areas of scale, speed, dynamic updating, condensed representation of the graph, and measure direct interaction between nodes.

Some forms of unsupervised neural networks have been applied. Dorronsoro *et al* (1997) creates a non-linear discriminant analysis algorithm which do not need labels. It minimises the ratio of the determinants of the within and between class variances of weight projections. There is no history on each credit card account's past transactions, so all transactions have to be segregated into different geographical locations. The authors explained that the installed detection system has low false positive rates, high cost savings, and high computational efficiency. Burge and Shawe-Taylor (2001) use a recurrent neural network to form short-term and long-term statistical account behaviour profiles. Hellinger distance is used to compare the two probability distributions and give a suspicion score on telecommunications toll tickets.

In addition to cluster analysis (Section 4.3.2), unsupervised approaches such as outlier detection, spike detection, and other forms of scoring have been applied. Yamanishi *et al* (2004) demonstrated the unsupervised SmartSifter algorithm which can handle both categorical and continuous variables, and detect statistical outliers using Hellinger distance, on medical insurance data.

Bolton and Hand (2001) recommend Peer Group Analysis to monitor inter-account behaviour over time. It compares the cumulative mean weekly amount between a target account and other similar accounts (peer group) at subsequent time points. The distance metric/suspicion score is a *t*-statistic which determines the standardised distance from the centroid of the peer group. The time window to calculate peer group is thirteen weeks and future time window is four weeks on credit card accounts. Bolton and Hand (2001) also suggest Break Point Analysis to monitor intra-account behaviour over time. It detects rapid spending or sharp increases in weekly spending within a single account. Accounts are ranked by the *t*-test. The fixed-length moving transaction window contains twenty-four transactions: first twenty for training and next four for evaluation on credit card accounts.

Brockett *et al* (2002) recommends Principal Component Analysis of RIDIT scores for rank-ordered categorical attributes on automobile insurance data.

Hollmen and Tresp (1998) present an experimental real-time fraud detection system based on a Hidden Markov Model (HMM).

## 4.6 Critique of Methods and Techniques

- In most scenarios of real-world fraud detection, the choice of data mining techniques is more dependent on the practical issues of operational requirements, resource constraints, and management commitment towards reduction of fraud than the technical issues poised by the data.

- Other novel commercial fraud detection techniques include graph-theoretic anomaly detection[2] and Inductive Logic Programming[3]. There has not been any empirical evaluation of commercial data mining tools for fraud detection since Abbott *et al* (1998).

- Only seven studies claim to be implemented (or had been) as actual fraud detection systems: in insurance (Major and Riedinger, 2002; Cox, 1995), in credit card (Dorronsoro *et al*, 1997; Ghosh and Reilly, 1994), and in telecommunications (Cortes *et al*, 2003; Cahill *et al*, 2002; Cox, 1997). Few fraud detection studies which explicitly utilise temporal information and virtually none use spatial information.

- There is too much emphasis by research on complex, non-linear supervised algorithms such as neural networks and support vector machines. In the long term, less complex and faster algorithms such as naive Bayes (Viaene *et al*, 2002) and logistic regression (Lim *et al*, 2000) will produce equal, if not better results (see Section 3.2), on population-drifting, concept-drifting, adversarial-ridden data. If the incoming data stream has to be processed immediately in an event-driven system or labels are not readily available, then semi-supervised and unsupervised approaches are the only data mining options.

- Other related data mining techniques covered by survey papers and bibliographies include outlier detection (Hodge and Austin, 2004), skewed/imbalanced/rare classes[4] (Weiss, 2004), sampling (Domingos *et al*, 2002), cost sensitive learning[5], stream mining[6], graph mining (Washio and Motoda, 2003), and scalability (Provost and Kolluri, 1999).

# 5. OTHER ADVERSARIAL DOMAINS

This section explains the relationship between fraud detection three other similar domains.

## 5.1 Terrorist Detection

There had been simplistic technical critiques of data mining for terrorist detection such as low accuracy (unacceptably high false positive rates in skewed data) and serious privacy violations (massive information requirements). To counter them, Jensen *et al* (2003) recommend fixed-size clustering to generate true class labels and the linked structure of data. Scores are randomly drawn from either the negative or positive entities' normal distributions. The second-round classifier averages an entity's first-round score and scores of all its neighbours. To reduce false positives, results show that second-round classifier reduces false positive rates while maintaining true positive rates of first-round classifier. To reduce information requirements, results show moderately high accuracy through the use of only twenty percent of the data.

Surveillance systems for terrorist, bio-terrorist, and chemo-terrorist detection often depend on spatial and spatio-temporal data. These are unsupervised techniques highly applicable to fraud detection. Neill and Moore (2004) employ Kulldorff's spatial scan statistic and the overlap-kd tree data structure. It efficiently finds the most significant densities from latitude and longitude of patient's home in real emergency department, and zip codes in retail cough and cold medication sales data. Das *et al* (2004) utilise Partially Observable Markov Decision Process (POMDP) with Kulldorff's spatial scan statistic on to detect artificial attacks from real emergency department's spatio-temporal data.

Bio-terrorism detection aims to detect irregularities in temporal data. Similar to fraud detection, data has to be partially simulated by injecting epidemics, and performance is evaluated with detection time and number of false positives. Wong *et al* (2003) apply Bayesian networks to uncover simulated anthrax attacks from real emergency department data. Hutwagner *et al* (2003) describe the use of cumulative sum of deviations in the Early Aberration Reporting System (EARS). Goldenberg *et al* (2002) use time series analysis to track early symptoms of synthetic anthrax outbreaks from daily sales of retail medication (throat, cough, and nasal) and some grocery items (facial tissues, orange juice, and soup). Other epidemic detection papers include application of sliding linear regression to usage logs of doctors' reference database (Heino and Toivonen, 2003) and HMMs to influenza time series (Rath *et al*, 2003).

## 5.2 Financial Crime Detection

Financial crime here refers to money laundering, violative trading, and insider trading and the following are brief application descriptions which correspond to each type of monitoring system for the United States government. The Financial Crimes Enforcement Network AI System (FAIS) (Senator *et al*, 1995) operates with an expert system with Bayesian inference engine to output suspicion scores and with link analysis to visually examine selected subjects or accounts. Supervised techniques such as case-based reasoning, nearest neighbour retrieval, and decision trees were seldom used due to propositional approaches, lack of clearly labelled positive examples, and scalability issues. Unsupervised techniques were avoided due to difficulties in deriving appropriate attributes. It has enabled effectiveness in manual investigations and gained insights in policy decisions for money laundering.

The National Association of Securities Dealers' (NASD) Regulation Advanced Detection System (ADS) (Kirkland *et al*, 1999) uses a rule pattern matcher and a sequence matcher cast in two- and three- dimensional visualisations to generate breaks or leads. The rule pattern matcher detects predefined suspicious behaviours; whilst the sequence matcher finds temporal relationships between events from market data which exists in a potential violation pattern. Association rules and decision trees are used to discover new patterns or refined rules which reflect behavioural changes in the marketplace. It has been successfully used to identify and correct potential violative trading on the NASDAQ National Market. Senator (2000) argues that propositional data mining approaches are not useful for the ADS.

The Securities Observation, News Analysis, and Regulation (SONAR) (Goldberg *et al*, 2003) uses text mining, statistical regression, rule-based inference, uncertainty, and fuzzy matching. It mines for explicit and implicit relationships among the entities and events, all of which form episodes or scenarios with specific identifiers. It has been reported to be successful in generating breaks the main stock markets for insider trading (trading upon inside information of a material nature) and misrepresentation fraud (falsified news).

Use of large amounts of unstructured text and web data such as free-text documents, web pages, emails, and SMS messages, is common in adversarial domains but still unexplored in fraud detection literature. Zhang *et al* (2003) presents Link Discovery on Correlation Analysis (LDCA) which uses a correlation measure with fuzzy logic to determine similarity of patterns between thousands of paired textual items which have no explicit links. It comprises of link hypothesis, link generation, and link identification based on financial transaction timeline analysis to generate community models for the prosecution of money laundering criminals.

Use of new relevant sources of data which can decrease detection time is another trend in adversarial domains which is lacking in fraud detection research. Donoho (2004) explores the use of C4.5 decision tree, backwards stepwise logistic regression, neural networks (all learning algorithms with default parameters), expert systems, and *k*-means clustering (only on positive instances). It is for finding early symptoms of insider trading in option markets before any news release.

## 5.3 Intrusion and Spam Detection

There are multiple data sources for intrusion detection and the common ones are at host level, network level, and user level. Otey *et al* (2003) advocates intrusion detection at the secure Network Interface Cards (NIC) level. Leckie and Yasinsac (2004) argue that in an encrypted environment, intrusion detection can utilise metadata at the user level. Julisch and Dacier (2002) mine historical alarms to reduce false positives. Shavlik and Shavlik (2004) monitor Windows® operating systems. Compared to fraud detection where many studies use real proprietary data, the benchmark KDD cup 1999 network intrusion detection data is often used. In addition, semi-real user level data are common, the "intrusions" are usually simulated using another user data and "real" part refers to normal computer usage data for each legitimate user.

In intrusion detection terms, misuse detection is for matching known attacks (using A of Figure 4.1); and anomaly detection is for discovering unknown attacks (using C of Figure 4.1 and see Section 4.4). The current research in both intrusion detection and spam detection are on anomaly detection (semi-supervised) and unsupervised approaches. In intrusion detection research, the use of clustering to reduce data and HMMs for anomaly detection had been popular. Lane and Brodley (2003) detail that *k*-means to compress data and report that HMMs performed slightly better than instance-based learning (IBL) for semi-real user level data. Similarly, Cho (1999) use SOM to decrease data for HMM modelling. The author show that multiple HMM models with fuzzy logic can be used to reduce false positive rates. Also, Stolfo *et al* (2001) advocate Sparse Markov Transducers. However, Yeung and Ding (2002) conclude that simple static approaches, such as occurrence frequency distributions and cross entropy between distributions, outperform HMMs.

Other anomaly detection studies trialled with Winnow (Shavlik and Shavlik, 2004), RIPPER (Fan *et al*, 2001), *Apriori* (Lee *et al*, 2000), frequent episodes (Julisch and Dacier, 2002; Lee *et al*, 2000), attribute-oriented induction (Julisch and Dacier, 2002), and *k*-means (Sequeira and Zaki, 2002). Most studies conclude that anomaly detection does not perform as well as misuse detection. Unsupervised approaches include Hawkins *et al* (2002) and Williams *et al* (2002) which advocate replicator neural networks to detect outliers. To detect spam from the email server, Yoshida *et al* (2004) encodes email as hash-based text, and avoids labels through the use of document space density to find large volumes of similar emails. On a small sample of labelled emails, the authors comment that SVM is the best supervised algorithm but the detection time is too long for an event-driven system.

The use of game theory to model the strategic interaction between the system and adversary has been recently introduced into intrusion and spam detection research. Patcha and Park (2004) apply theoretical game theory to account for the interaction between one attacker and one node to detect intrusions from mobile ad hoc networks. In spam detection, Dalvi *et al* (2004) adapt game theory to automatically re-learn a cost-sensitive supervised algorithm given the cost-sensitive adversary's optimal strategy. It defines the adversary and classifier optimal strategy by making some valid assumptions. Tested under different false positives costs, the game-theoretic naive Bayes classifier outperforms the conventional classifier by efficiently predicting no false positives with relatively low false negatives.

## 6. RELATED WORK

Bolton and Hand (2002) discuss techniques used in several subgroups within fraud detection such as credit card and telecommunications, and related domains such as money laundering and intrusion detection. Kou *et al* (2004) outline techniques from credit card, telecommunications, and intrusion detection. Weatherford (2002) recommends backpropagation neural networks, recurrent neural networks and artificial immune systems for fraud detection.

This paper examines fraud detection from a practical data-oriented, performance-driven perspective rather than the typical application-oriented or technique-oriented view of the three other recent survey papers.

In addition, this survey clearly defines the underlying technical problems and covers more relevant fraud types, methods, and techniques than any of the other survey papers. For example, internal fraud and the various hybrid approaches are presented here. Also, some criticisms of the current fraud detection field are given and possible future contributions to data mining-based fraud detection from related domains are highlighted.

## 7. CONCLUSION & FUTURE WORK

This survey has explored almost all published fraud detection studies. It defines the adversary, the types and subtypes of fraud, the technical nature of data, performance metrics, and the methods and techniques. After identifying the limitations in methods and techniques of fraud detection, this paper shows that this field can benefit from other related fields. Specifically, unsupervised approaches from counterterrorism work, actual monitoring systems and text mining from law enforcement, and semi-supervised and game-theoretic approaches from intrusion and spam detection communities can contribute to future fraud detection research. However, Fawcett and Provost (1999) show that there are no guarantees when they successfully applied their fraud detection method to news story monitoring but unsuccessfully to intrusion detection. Future work will be in the form of credit application fraud detection.

## ACKNOWLEDGEMENTS

This research is financially supported by the Australian Research Council, Baycorp Advantage, and Monash University under Linkage Grant Number LP0454077. The authors gratefully acknowledge the detailed comments from Dr. Wong Weng-Keen and insights provided by Prof. John Galloway.

## NOTES

[1] http://www.kdnuggets.com

[2] http://www.idanalytics.com/solutions/gtad-technology.html

[3] http://www.futureroute.co.uk/pressreleaseaug17.html

[4] http://www.site.uottawa.ca/~nat/Research/class_imbalance_bibli.html

[5] http://members.rogers.com/peter.turney/bibliographies/cost-sensitive.html

[6] http://www.csse.monash.edu.au/~mgaber/WResources.htm

[7] http://liinwww.ira.uka.de/bibliography/Ai/fraud.detection.html

## REFERENCES[7]